\documentclass[conference]{IEEEtran}
\usepackage{times}

\usepackage[numbers]{natbib}
\usepackage{multicol}
\usepackage[bookmarks=true]{hyperref}
\usepackage{amsmath}
\usepackage{multirow}   
\usepackage{amssymb}
\usepackage{graphicx}
\usepackage{caption} 
\usepackage{booktabs}
\usepackage{bm}
\usepackage{xcolor}
\usepackage{float}
\usepackage{changepage}

\IEEEoverridecommandlockouts 

\IEEEaftertitletext{%
  \begin{center}
    \vspace{-1em} 
    
    \includegraphics[width=1\textwidth]{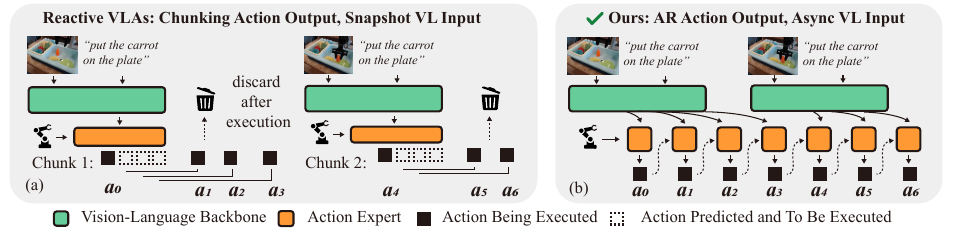}
    
    \captionof{figure}{(a) The prevalent approach in Vision-Language-Action models predicts action chunks based only on the current snapshot of information. It discards the temporal context and drops the state update within an action chunk execution, leading to reactive, memoryless prediction. (b) In contrast, we propose AR-VLA that leverages an autoregressive action expert that maintains its own history through a
long-lived memory with inherent context-awareness. Visual-language conditions are updated asynchronously without interrupting the action stream.}
    \label{fig:vs}

    \vspace{1em} 
  \end{center}
}

\begin{document}

\title{AR-VLA: Autoregressive Action Expert for Vision–Language–Action Models}




%

\author{\authorblockN{Yutong Hu\authorrefmark{1}\authorrefmark{2}\authorrefmark{4}
Jan-Nico Zaech\authorrefmark{1}
Nikolay Nikolov\authorrefmark{1}
Yuanqi Yao\authorrefmark{1}
Sombit Dey\authorrefmark{1} Giuliano Albanese\authorrefmark{1} \\ Renaud Detry\authorrefmark{2}\authorrefmark{3}\authorrefmark{4} Luc Van Gool\authorrefmark{1} Danda Paudel\authorrefmark{1}}
\authorblockA{\authorrefmark{1}INSAIT, Sofia University “St. Kliment Ohridski”}
\authorblockA{\authorrefmark{2}KU Leuven, Dept. Mechanical Engineering, Research unit Robotics, Automation and Mechatronics}
\authorblockA{\authorrefmark{3}KU Leuven, Dept. Electrical Engineering, Research unit Processing Speech and Images}
\authorblockA{\authorrefmark{4}Flanders Make@KU Leuven}

}

\maketitle

\begin{abstract}

We propose a standalone autoregressive (AR) Action Expert that generates actions as a continuous causal sequence while conditioning on refreshable vision-language prefixes. In contrast to existing Vision-Language-Action (VLA) models and diffusion policies that reset temporal context with each new observation and predict actions reactively, our Action Expert maintains its own history through a long-lived memory and is inherently context-aware. This structure addresses the frequency mismatch between fast control and slow reasoning, enabling efficient independent pretraining of kinematic syntax and modular integration with heavy perception backbones, naturally ensuring spatio-temporally consistent action generation across frames. To synchronize these asynchronous hybrid V-L-A modalities, we utilize a re-anchoring mechanism that mathematically accounts for perception staleness during both training and inference. Experiments on simulated and real-robot manipulation tasks demonstrate that the proposed method can effectively replace traditional chunk-based action heads for both specialist and generalist policies. AR-VLA exhibits superior history awareness and substantially smoother action trajectories while maintaining or exceeding the task success rates of state-of-the-art reactive VLAs. Overall, our work introduces a scalable, context-aware action generation schema that provides a robust structural foundation for training effective robotic policies. Code and Videos available at \url{https://arvla.insait.ai/}
\end{abstract}

\IEEEpeerreviewmaketitle

\begin{figure*}[t]
    \centering
    \includegraphics[width=1\linewidth]{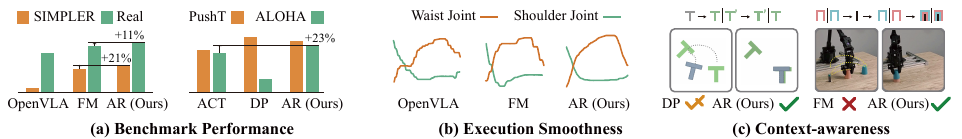}
    \caption{\textbf{Performance Overview.} (a) Quantitative Results: In both generalist (left) and specialist (right) benchmarks, AR-VLA achieves competitive or superior performance compared to state-of-the-art policies, including OpenVLA, Flow-Matching (FM), ACT, and Diffusion Policy (DP), details in Sec.\ref{sec:bench}. (b) Trajectory Quality: Qualitative visualization of joint trajectories over time reveals that AR-VLA produces significantly smoother and more kinematically consistent motion compared to reactive baselines that reset context at each step (analysis in Sec.\ref{sec:smoothness}). (c) Long-Horizon Capability: AR-VLA successfully completes long-horizon tasks where baselines like DP and FM fail due to a lack of temporal context awareness. Detailed task defination and explanation in Sec. \ref{sec:history}.}
    \label{fig:teaser}
\end{figure*}
\section{Introduction}
The ``next-token prediction" paradigm has emerged as one of the primary engines of modern artificial intelligence. Large-scale autoregressive models, such as LLMs~\cite{llm} and VLMs~\cite{beyerPaligemmaVersatile3b2024}, demonstrate that the synergy of causal sequence modeling, scalable attention, and massive computation is essential for the appearance of emergent reasoning and robust generalization. Naturally, this paradigm is now being extended from sequences of words to sequences of actions via Vision-Language-Action (VLA) models. However, while recent VLA architectures (e.g., OpenVLA~\cite{kimOpenvlaOpensourceVisionlanguageaction2024b}, RT-2~\cite{zitkovichRt2VisionlanguageactionModels2023}, Pi-0-FAST~\cite{pertschFastEfficientAction2025}) are frequently labeled ``autoregressive", this terminology is deceptive in the context of robotic control. These models utilize autoregression only to generate tokens within a single inference step. Effectively, they do not autoregress across time. 

Current state-of-the-art robot learning methods, including Diffusion Policies~\cite{chiDiffusionPolicyVisuomotor2023a} and existing VLAs, treat action generation not as a continuous stream, but as a series of isolated events. As shown in Fig.~\ref{fig:vs}(a), these models typically employ ``action chunking"~\cite{zhaoLearningFineGrainedBimanual2023c}: predicting a static block of actions at once, directly or through iterative denoising. While effective for short-horizon smoothness, these approaches remain structurally reactive: at every perception step, the model acts as if it is ``waking up'' for the first time, re-encoding the visual context and generating a trajectory chunk without a persistent internal state of its own perception and action history. Consequently, they suffer from ``Markovian amnesia", discarding temporal continuity and degrading fluid control to a series of disjointed, snapshot-conditioned responses.


We argue that manipulation is not merely a stack of separate visual-motor snapshots; it is a problem of streaming control. To act effectively, a policy requires two distinct forms of awareness: \textbf{situational} awareness (semantic understanding of ``what" is in the workspace and ``where" the robot is) and \textbf{temporal} awareness (kinematic understanding of ``what" has already occurred and ``how" the end-effector is accelerating). While VLMs provide the former, they are structurally ill-suited for the latter due to their high latency and episodic nature. The missing piece here is a truly Autoregressive Action Expert -- just as an LLM predicts the next word based on the ``flow" of a conversation, a robot policy should predict the next pose based on the ``momentum" of its trajectory.

By treating action as a ``language of motion", a true Autoregressive Action Expert provides three transformative benefits to the VLA paradigm, as in Fig.~\ref{fig:teaser}. \textbf{(1) It is naturally context-aware,} as its internal state captures the causal dependencies of the entire trajectory rather than reacting to a local snapshot. \textbf{(2) It is naturally decoupled from the VLM backbone,} allowing the motor thread to run at high frequencies with temporal consistency, regardless of perception latency. \textbf{(3) It facilitates independent pretraining using only the action labels}, enabling the model to master the syntax of movement (dynamics, joint constraints, and physical causality) on large-scale kinematic data before the visual alignment phase.

To realize these potentials, we introduce AR-VLA, a unified framework that instantiates such an action expert within a single architecture for both robot specialists and generalists. As in Fig.~\ref{fig:vs}(b), AR-VLA structurally decouples the high-level semantic reasoning of vision-language models from the high-frequency temporal consistency of robot control. Rather than treating the action head as a dependent appendage of a VLM, we formulate it as an independent expert that maintains a continuous, evolving memory of its own history. This design preserves long-horizon intent while allowing the model to asynchronously attend to the latest visual-language features provided by a VLM. This architecture bridges a fundamental frequency mismatch in robotics, providing a solution one step closer to the \textbf{system 1/2}~\cite{kahnemanThinkingFastSlow2011, ahnCanNotSay2022} dichotomy: the ``brain" (semantic perception) updates slowly, while the ``cerebellum" (motor control) streams high-frequency commands.

Our core contribution is\textbf{ the formulation of an Autoregressive Action Expert}, which treats action generation as a causal sequence modeling problem across time. By maintaining a long-lived context of past actions, our model inherently resolves the temporal inconsistency and ``jitter" prevalent in reactive policies, outperforming denoise/chunk-based baselines in trajectory smoothness and long-horizon stability. To instantiate this expert, we propose two technical pillars: \textbf{(1) Hybrid Key-Value (HKV) Cache:} A novel Transformer decoder architecture that manages two distinct memory streams: a rolling, token-wise FIFO for high-frequency actions and a block-wise, refreshable buffer for low-frequency visual semantics. This allows the action stream to function as an independent expert that is ``guided" rather than ``blocked" by perception. \textbf{(2) Dynamic Temporal Re-anchoring (DTR):} We solve the synchronization challenge of asynchronous streams via DTR, a mechanism that explicitly anchors visual keys based on their capture-time index. This ensures the model mathematically understands the ``staleness" of a visual frame, bridging the gap between short-context training and long-horizon inference.


\section{Related Work}
\label{sec:related_work}

\subsection{Vision-Language-Action models (VLAs).}
Traditional robot imitation learning \citep{shafiullahBehaviorTransformersCloning2022, cuiPlayPolicyConditional2022, chiDiffusionPolicyVisuomotor2023a, leeBehaviorGenerationLatent2024a, shridhar2022peract} has typically relied on task-specific data, which covers only narrow distributions of environments and instructions. To mitigate this, recent research \citep{zitkovichRt2VisionlanguageactionModels2023, driessPalmeEmbodiedMultimodal2023, kimOpenvlaOpensourceVisionlanguageaction2024b, pertschFastEfficientAction2025, bjorckGr00tN1Open2025, spiridonov2025generalist, blackPi_0VisionlanguageactionFlow2025, goyal2024rvt, zawalski2024robotic, reuss2025FLOWER, yang2025instruct, kawaharazuka2025vla-survey} suggests constructing generalist policies by injecting internet-scale VLM priors into low-level action generation, effectively transferring broad semantic understanding to physical control. Initial VLA models~\citep{zitkovichRt2VisionlanguageactionModels2023, kimOpenvlaOpensourceVisionlanguageaction2024b, dey2024revlarevertingvisualdomain, driessPalmeEmbodiedMultimodal2023, kim2025fine, rt12022arxiv} often discretized action spaces to treat control as a token prediction task. Subsequent developments \citep{blackPi_0VisionlanguageactionFlow2025, bjorckGr00tN1Open2025, liCogactFoundationalVisionlanguageaction2024, black2025real, song2025hume} have utilized VLM embeddings to condition continuous action generators via diffusion or flow-matching. For example, $\pi_0$~\citep{blackPi_0VisionlanguageactionFlow2025} conditions a flow-matching head on VLM features to generate multi-step action chunks, while CogAct \citep{liCogactFoundationalVisionlanguageaction2024} demonstrates the scalability of diffusion action transformers when grounded in VLM representations. However, these architectures remain predominantly reactive, basing control decisions on the immediate observation and often ignoring the temporal context vital for accurate state estimation. In this work, we focus on enhancing current VLAs by integrating persistent historical context into the autoregressive process, without changing any of the Vision-Language perception part.

\subsection{Action Representation and Pretraining.}
Ensuring that features from different domains reside in well-structured representation spaces is foundational for effective cross-modality alignment. In the action domain, given its relatively low dimensionality, this is traditionally achieved through classical methods such as statistical normalization~\citep{blackPi_0VisionlanguageactionFlow2025}, categorical binning~\citep{kimOpenvlaOpensourceVisionlanguageaction2024b}, or discretization via $k$-means clustering~\citep{shafiullahBehaviorTransformersCloning2022}. Distinct from these approaches, there is a growing interest in modern representation learning that treats actions as a temporal sequence. By leveraging large-scale trajectory datasets prior to visual conditioning, these models learn to capture the fundamental motion primitives and dynamics of the embodiment. Recent advances in action tokenization demonstrate that robust motion priors can be learned efficiently through explicit modeling, such as Fast~\citep{pertschFastEfficientAction2025}, FASTer~\citep{liuFASTerEfficientAutoregressive2025}, BEAST~\citep{zhou2025beast}, and OmniSAT~\citep{lyu2025omnisatcompactactiontoken} or implicit neural architectures like Vector Quantized Variational Autoencoders (VQ-VAE)~\citep{leeBehaviorGenerationLatent2024a, ye2025latent}. These approaches successfully capture the underlying syntax of robot kinematics to facilitate the final action prediction. Our approach extends this idea further by treating the pretrained action model not merely as a passive token translator, but as a standalone autoregressive expert. This formulation enables the model to perform implicit sequence modeling of motion priors while simultaneously allowing for asynchronous coupling with high-latency, heavy perception modules.

\subsection{Architectures with context awareness}
While many current robotic datasets and benchmarks are limited to short-horizon tasks, most real-world applications are inherently long-horizon and non-Markovian, necessitating memory mechanisms that allow policies to utilize historical observations and actions. In reinforcement learning, recurrent policies \citep{hausknechtDeepRecurrentQlearning2015} and Transformer-based variants \citep{parisottoStabilizingTransformersReinforcement2020} established memory as a primary tool for performance in partially observable environments. Similarly, modern natural language models leverage KV-caching to remain aware of historical context. Explicit memory structures, ranging from end-to-end memory networks \citep{sukhbaatarEndtoendMemoryNetworks2015} to retrieval-based models \citep{khandelwalGeneralizationMemorizationNearest2019, lewisRetrievalaugmentedGenerationKnowledgeintensive2020}, have been thoroughly investigated to bolster long-context reasoning. Despite these developments, memory mechanism in VLA models remains under-explored. While most VLAs are context-unaware, MemoryVLA \citep{shiMemoryvlaPerceptualcognitiveMemory2025} proposes architectures inspired by human cognitive systems but requires training from scratch. HAMLET \citep{myungkyukoodaewonchoitaeyoungkimkyungminleechangyeonkimyounggyoseojinwooshinHAMLETSwitchYour2025} augments pretrained VLAs with learnable tokens and memory modules to achieve history-awareness without full retraining. Distinct from these, our approach utilizes a true autoregressive model for the action expert, making it naturally holds a record of the system's evolution from historical states and providing an innate context awareness.

\section{Methodology}

The AR-VLA framework bridges the divide between high-latency semantic perception and high-frequency motor control through a stateful, two-stage architecture. At its core is a standalone autoregressive action expert that maintains kinematic continuity via a \textbf{Hybrid Key-Value Cache (HKV)}, allowing it to marry a rolling proprioceptive history with a refreshable visual-linguistic context. To align these asynchronous streams, we introduce \textbf{Dynamic Temporal Re-anchoring (DTR)}, a position-encoding mechanism that maps static visual-language features onto the dynamic action timeline. Our training protocol unfolds in two phases: (1) action-only pretraining to master the syntax of motion, and (2) cross-modal alignment to ground that motion in visual perception.

\subsection{Problem Formulation}

We formalize a robot trajectory $\{\tau\}$ as a sequence of observations and actions $\{ (o_t, a_t)_{t=0}^T \}$. The observation $o_t$ is decomposed into exteroceptive inputs (visual frames $v_t$ and language instructions $l$) and proprioceptive states $s_t$.

A common workaround to circumvent the limitations of stateless architectures involves stacking observations temporally and predicting actions in chunks. While this constructs a ``pseudo-history,'' it still fails to cure the underlying Markovian bottleneck; the model have to re-infer history intent and even current velocity from scratch at every new window, often resulting in jittery control and temporal incoherence.

\smallskip
\noindent\textbf{Definition 1: The Reactive Actor.} Standard VLAs typically map the \textit{current} observation to the \textit{current} action, resetting their context memory at each step $t$: 
\begin{equation}
P_{\text{react}}(\tau) = \prod_{t=1}^T P(a_t \mid \Phi(v_t, l), s_t),
\end{equation}
where $\Phi(\cdot)$ denotes the perception encoder (e.g., a VLM or a smaller backbone) used to embed the observations.

\smallskip
\noindent\textbf{Definition 2: The Autoregressive (AR) Actor.} We define the AR Actor as a sequence model where one of the prediction dependencies is the continuous kinematic history, while remaining conditioned on the most recently available visual-language prefix from $\Phi$. Let $i \le t$ be the index of the most recent visual frame processed:
\begin{equation}
P_{\text{ar}}(\tau) = \prod_{t=1}^T P(a_t \mid \underbrace{\Phi(v_i, l)}_{\text{VL Prefix}}, \underbrace{a_{<t}, s_{<t}}_{\text{Kinematic History}}).
\end{equation}
In this formulation, $a_t$ depends explicitly on the continuous causal chain $a_{<t}, s_{<t}$. This persistent memory ensures kinematic smoothness and robustness to visual-language (VL) latency.

\subsection{Model Structure}

AR-VLA is instantiated as a unified Transformer decoder designed to process a hybrid stream of continuous proprioceptive data and high-dimensional VL embeddings.

    \smallskip
\noindent\textbf{Continuous Action Representation.} To preserve the precision required for low-level manipulation, we follow the convention of continuous action regression. Yet the AR training naturally fits discrete token supervision. Robot actions $a_t \in \mathbb{R}^d$ represent target end-effector pose deltas or joint velocities. Within the Transformer, these vectors are projected to the model dimension $D$ via a linear layer, treating each timestep's vector as a single token. The model output is regressed to $a_{t+1}$ via a deterministic prediction head. Because $s$ and $a$ vary across robots and tasks, we denote arbitrary action-modality tokens as $x$ to maintain a consistent autoregressive notation. In the following part, we use $X$ to mark the proprioceptive stream and $VL$ to mark the language stream.

\begin{figure}
    \centering
    \includegraphics[width=1\linewidth]{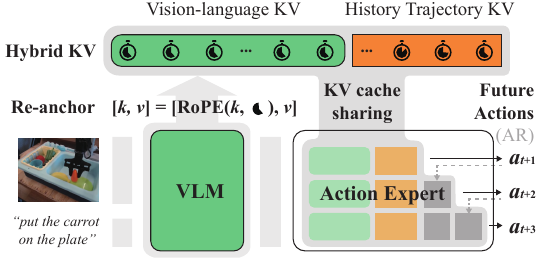}
    \caption{\textbf{The AR-VLA Framework}. The system bridges an VLM backbone with a autoregressive Action Expert asynchronously. Atemporal features from the VLM are explicitly injected with temporal context via Dynamic Temporal Re-anchoring (DTR). Within the Hybrid KV Cache, re-anchored VL tokens (green) serve as a semantic prefix to the rolling kinematic history (orange). The Action Expert generates future action sequences by querying this shared cache using incrementally advancing step embeddings.}
    \label{fig:framework}
\end{figure}

    \smallskip
\noindent\textbf{Unified Decoder with Hybrid Cache.} The architecture relies on a Hybrid Key-Value (HKV) Cache that manages two heterogeneous sources of context. As in Fig.\ref{fig:framework}, we apply distinct update rules, enabling the structural decoupling of perception and control: {(1) {Proprioceptive Stream ($KV^X$):}} A rolling FIFO buffer storing the KV pairs of the robot's state and action history. This long-lived window is significantly longer than the history stacks (1$\sim$4) used in reactive VLAs, capturing the momentum necessary for stability. {(2) {Visual-Language Stream ($KV^{VL}$):}} A single-slot buffer storing KV pairs projected from the VLM backbone. This acts as a refreshable semantic prefix, replaced entirely whenever a new frame is processed.

    \smallskip
\noindent\textbf{Dynamic Temporal Reanchoring (DTR).} The fundamental challenge in decoupling these threads is temporal alignment. We introduce DTR, which leverages the mathematical properties of Rotary Positional Embeddings (RoPE \cite{su2023roformerenhancedtransformerrotary}) to encode the relative distance between the high-frequency action stream and the inherently atemporal VL context retrieved from the VLM.

In our unified decoder, the attention output for an action query at temporal index $m$ is calculated over the combined set of cached proprioceptive and VL key-value pairs:
\begin{equation}
\text{Attn}(q_m, \bm{K}, \bm{V}) = \sum_n \text{softmax} \left( \frac{\text{Score}(q_m, k_n)}{\sqrt{d}} \right) v_n, 
\end{equation}
where $\text{Score}(q_m, k_n)$ denotes the position-aware inner product. RoPE implements this by applying a rotation matrix $\bm{R}(p)$ to the feature vectors at position $p$, such that the attention score depends only on the relative distance. Crucially, because VLM embeddings are generated independently of the robot's trajectory, we manually assign indices to bridge the training-inference gap:
\begin{equation}
k_n = \bm{R}(n)\bm{k}, \quad v_n = \bm{v} ,
\end{equation}
where action tokens $\bm{k}^X$ and VL tokens $\bm{k}^{VL}$ are treated differently when assigning their anchor index $n$: 
(1) Action Tokens:  $\bm{k}^X$ follow the robot's causal timeline. They are assigned the sequential index $n$ corresponding to the timestep at which they were executed. 
(2) VL Tokens:  $\bm{k}^{VL}$ from VLM backbone are inherently atemporal. Without extra processing,
those tokens only form a static semantic snapshot unaware of
the robot’s current step. To encode their temporal validity, we assign them the fixed index $n$ corresponding to the timestep when the image was captured. 

By defining $n$ in this way, the relative distance $(m - n)$ between the current query $m$ and the VL key $n$ mathematically represents the data staleness, as in Fig.\ref{fig:framework}. The resulting interaction becomes $\text{Score}(q_m, k^{VL}_n) = \bm{q}^\top \bm{R}(m - n) \bm{k}^{VL}$. A vital property of this formulation is that the score remains identical under a global time shift $T$:
\begin{equation}
\text{Score}(q_{m+T}, k^{VL}_{n+T}) =\text{Score}(q_m, k^{VL}_n) .
\end{equation}

Because the VL values $v^{VL}$ remain constant and un-rotated, the resulting weighted sum in the attention mechanism is purely a function of the relative staleness $\Delta t = m - n$.

This mechanism is critical for resolving the discrepancy between training and deployment. During training, we typically sample short batches (e.g., current step $m=25$ and an image anchored at $n=20$). In this scenario, the model learns to act based on a VL context with a staleness of $\Delta t = 5$. During real-world inference, the robot may reach global step 500 while still processing a VL update from step 495. Though the staleness of $\Delta t = 5$ remains in-distribution, the absolute indices $(500, 495)$ would fall far outside the distribution seen during training if we are not using DTR with RoPE, leading to unpredictable behavior. By exploiting the shift-invariance property ($T=475$), DTR ensures that $\text{Score}(q_{500}, k^{VL}_{495}) = \text{Score}(q_{25}, k^{VL}_{20})$. This allows the model to apply the same visual grounding logic whether a seen situation occurs at step 25 or step 500.

\subsection{Training Details}

Our training protocol follows a two-phase regime to master motion syntax before grounding it in perception.

\smallskip
\noindent\textbf{Phase 1: Action-Only Pretraining}: The actor is first optimized on large-scale trajectories as a standalone autoregressive action sequence generator. Using a causal mask and sequential RoPE indices, we optimize the sequence modeling objective
\begin{equation}
\mathcal{L}_{\text{Phase1}} = \sum_{t=1}^{T} \mathcal{L}(x_t \mid x_{<t}). 
\end{equation}

This establishes a ``proprioceptive expert'' that masters kinematic syntax (e.g., joint limits, profiles, common move patterns) independently of VL data.

\smallskip
\noindent\textbf{Phase 2: VL-Action Alignment: }
We connect the VLM backbone to the expert using DTR, as in Fig.\ref{fig:framework}. Given a training sample $(\bm{x}_{\text{past}}, v, l, \bm{x}_{\text{fut}})$ where $v$ is an observation at time $H$, we do:
{(1) Priming:} History $\bm{x}_{\text{past}}$ is fed into the actor with indices $\{0, \dots, H-1\}$.
{(2) Anchoring: }VL features from $(v, l)$ are assigned the fixed index $H$, anchoring the perception to the history-future junction.
{(3) Stochastic Supervision with Historical Dropout:} The model predicts a horizon of $M$ future actions $\bm{x}_{\text{fut}}$ starting from the temporal anchor. To simulate execution noise and prevent parasitic over-reliance on history, we apply a unique random binary mask $\mathcal{M}_k \in \{0, 1\}^{H}$ for every individual future token $k \in \{0, \dots, M-1\}$. This forces the model to attend to the $VL$ prefix when historical context is corrupted or missing. The final loss is formulated as,
\begin{equation}
\mathcal{L}_{\text{Phase2}}=\sum_{k=0}^{M-1}\mathcal{L}(x_{H+k} \mid \mathcal{M}_k \odot \bm{x}_{\text{past}}, \Phi(v_H, l_H), \bm{x}_{H:H+k-1}). 
\end{equation}


\begin{figure}
    \centering
    \includegraphics[width=1.0\linewidth]{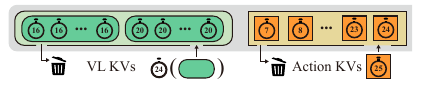}
    \caption{\textbf{Heterogeneous FIFO Update Rules for the Hybrid KV Cache.} The framework manages memory through two distinct queueing strategies to ensure efficient context utilization. The VL Stream (green) operates as a short-lived, block-wise FIFO: In contrast, the Action Stream (orange) maintains a token-wise rolling FIFO, continuously appending the single latest action prediction while evicting the oldest kinematic state.}
    \label{fig:fifo}
    \vspace{-10pt}
\end{figure}

\subsection{Inference Details}

Following training, the model effectively functions as a conditional next-token predictor $P(x_t \mid x_{<t}, \Phi(v_i,l_i))$, capable of generating precise actions even when the visual-language prefix $\Phi(v_i,l_i)$ is ``outdated'' relative to the current timestep $t$ ($i \le t$). This capability stems from the DTR mechanism, which ensures the attention mechanism generalizes to varying temporal offsets $\Delta t = t - i$, and the teacher-forcing training regime. By supervising the autoregressive prediction of a future horizon rather than a single step, the model learns to utilize visual-language prefixes anchored at various historical offsets (e.g., $t, t-1, \dots, t-k$), ensuring robustness to latency.

Consequently, the next action prediction relies purely on a hybrid KV cache comprising the up-to-date action domain history and  the most recently available visual-language tokens. This runtime Hybrid KV cache is constructed and maintained dynamically during inference, as in Fig. \ref{fig:fifo}. The action domain cache $KV^X$ operates as a persistent FIFO buffer, preserving the long-term history of the trajectory. Conversely, the visual-language cache $KV^{VL}$ is treated as a refreshable snapshot (functionally a single-slot FIFO buffer). Whenever a new visual embedding is available (captured at time $i$), we explicitly apply the DTR to the keys and replace the $KV^{VL}$ content, effectively ``snapping'' the perception to the new timestamp.

During deployment, such a hybrid composition of the KV cache allows the VLM and the Action Expert of AR-VLA execute their forward loop asynchronously. Therefore, AR-VLA supports both serial and parallel execution modes via a decoupled dual-thread architecture: \textbf{(1) Action Thread:} operates at a high control frequency. It autoregressively generates actions $a_t$, updates the $KV^X$ cache, and increments the global time index $t$. \textbf{(2) Perception Thread:} operates at the native frequency of the VLM. It processes the latest frame $v$ and asynchronously pushes updates to the $KV^{VL}$ buffer.

This structural decoupling provides a significant advantage over denoise-based chunking models. Even in serial execution, our lightweight action head offers lower latency per step than full denoising inference. In parallel execution, this design further guarantees a consistent control frequency, as the action prediction eliminates the blocking dependency on specific visual-language frames. The model simply conditions on the internal kinematic model ($KV^X$) and the latest valid VL prefix until a new one becomes available, ensuring a native smooth, uninterrupted actuation.

\section{Experiments}
\label{sec:experiments}
The objective of our experimental evaluation is to assess AR-VLA's capability for both specialist and generalist robot policies. Our comprehensive experiments aim to demonstrate that the autoregressive architecture provides a competitive alternative to existing action experts while offering superior history awareness and high-frequency control capabilities. The experiments are designed to answer the following questions:
\begin{enumerate}[label=\arabic*)]

    \item How well does AR Actor perform when replacing standard action heads in specialist and generalist policies? 
    \item How well does AR-VLA perform in terms of inference efficiency and trajectory quality?
    \item Can AR-VLA effectively solve long-horizon tasks that require history awareness?
    \item How do causal pretraining, RoPE anchoring, training-time history masking and inference-time AR context length contribute to the model's performance?
\end{enumerate}

\begin{figure}[t]
    \centering
    \includegraphics[width=1.0\linewidth]{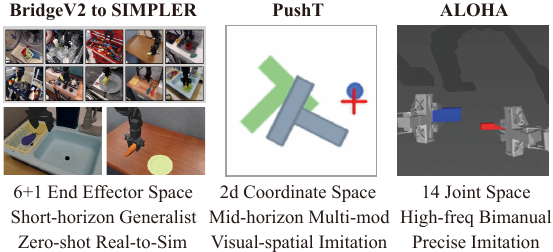}
    \caption{\textbf{Simulation benchmarks setups.} We do simulation evaluation spanning generalist and specialist policies, with diverse embodiment, action space, and task.}
    \label{fig:simulation_all}
    \vspace{-10pt}
\end{figure}

To answer these questions, as shown in Fig.~\ref{fig:simulation_all} and Fig.~\ref{fig:widowx}, we evaluate AR-VLA across both generalist VLA policies and specialist task-specific scenarios. 
Firstly, we assess AR-VLA's performance when replacing standard action heads by training on BridgeV2 and evaluating on SimplerEnv and real-world WidowX robot, comparing against existing VLA policies and baseline action experts with identical VLM backbones, as well as on three specialist manipulation tasks comparing with ACT and Diffusion Policy. 
Secondly, we quantify AR-VLA's efficiency advantages in inference frequency and trajectory smoothness.
Then, we design two tasks that explicitly require history awareness: PushT2 and Stack3, where critical information becomes unobservable without maintaining action history.  Finally, we conduct ablation studies to verify key design choices including causal pretraining and RoPE anchoring.

\subsection{Generalist and Specialist Policy Performance} \label{sec:bench}
\noindent\textbf{Evaluation Setups and Comparisons.}
We evaluate AR-VLA's capability to replace standard action prediction heads across generalist VLA and specialist policy settings.
For generalist VLA policies, we train on the BridgeV2 dataset~\citep{walkeBridgeDataV2Dataset2024} and create three models of identical 3B + 300M scale sharing the same Paligemma-3B \cite{beyerPaligemmaVersatile3b2024} VLM backbone with knowledge insulation~\citep{pertschFastEfficientAction2025} strategy: One predicts actions by decoding Fast tokens (i.e., a reproduced Pi-0-FAST*\citep{pertschFastEfficientAction2025}), one predicts action chunks through multi-step flow matching (Pi-0.5*\citep{intelligence$p_05$VisionLanguageActionModel2025a}), and one predicts actions autoregressively with standard next-action prediction loss (AR-VLA, Ours).
We evaluate on the SimplerEnv simulator~\citep{liEvaluatingRealWorldRobot2024a}, which provides WidowX manipulation tasks that replicate real-world robot conditions. We compare against OpenVLA~\citep{kimOpenvlaOpensourceVisionlanguageaction2024b}, Octo-Base, Octo-Small~\citep{teamOctoOpenSourceGeneralist2024a}, SpatialVLA~\citep{quSpatialVLAExploringSpatial2025a}, and CogACT~\citep{liCogACTFoundationalVisionLanguageAction2024a}.
For real-world evaluation, we deploy models on the WidowX robot following the BridgeV2 setup at 5Hz with the VLM refreshed every 4 actions. We establish a baseline where all polices succeed on all trials of the in-distribution task ``put eggplant into sink," then test on challenging tasks with 4 layout variants repeated 3 times.

For specialist policies, we evaluate AR Actor on PushT, ALOHA cube transfer, and ALOHA peg insertion using LeRobot~\citep{LeRobotOpenSourceLibrary2025}, comparing with Action Chunking Transformer (ACT~\citep{zhaoLearningFineGrainedBimanual2023c}) and Diffusion Policy (DP~\citep{chiDiffusionPolicyVisuomotor2023a}). To investigate AR Actor's capability to replace chunk-based action experts, we maintain ACT's exact encoder-decoder architecture and parameter size, but use it in a way following the same design principle as AR-VLA: we treat the encoder as a compact ``VLM" that caches intermediate key-values, while the decoder serves as an autoregressive action expert trained with next-token prediction loss. All methods share ResNet-18 vision encoder, and detailed model architecture in Appendix. 

\smallskip
\noindent\textbf{SimplerEnv Simulation Performance.}
Table~\ref{tab:simpler_vla} presents the evaluation results on SimplerEnv Visual Matching setting with four WidowX manipulation tasks: put spoon on towel, put carrot on plate, stack green block on yellow block, and put eggplant in yellow basket. On average, AR-VLA achieves the highest success rate of 61.5\%, significantly outperforming the second-best policy CogACT by a +9.4\% margin (52.1\%). Notably, despite sharing the identical VLM backbone with Pi-0-Fast and Pi-0.5, AR-VLA demonstrates superior action sequence generation capabilities through stored history key-value caches, surpassing Pi-0-Fast (49.0\%) and Pi-0.5 (51.0\%) by substantial margins.
The performance gains are particularly evident across individual tasks. AR-VLA achieves 75.0\% success on the spoon task, exceeding Pi-0-Fast (62.5\%) and Pi-0.5 (58.3\%). On the carrot task requiring precise manipulation, AR-VLA reaches 54.2\% success rate, substantially outperforming Pi-0-Fast (29.2\%) and Pi-0.5 (33.3\%). These results demonstrate that AR-VLA achieves better action sequence generation based on stored history key-value caches.
\begin{table}[t]
\centering
\caption{BridgeV2 Pretraining to SIMPLER Simulation Success Rate (\%)}
\begin{tabular}{l c c c c c}
\toprule
Model & Spoon & Carrot & Block & Eggplant & Average \\
\midrule
OpenVLA\citep{kimOpenvlaOpensourceVisionlanguageaction2024b}           & 0 & 0 & 0 & 4.1 & 1.0 \\
Octo-Base\citep{octomodelteamOctoOpensourceGeneralist2024}         & 12.5 & 8.3 & 0 & 43.1 & 16.0 \\
Octo-Small\citep{octomodelteamOctoOpensourceGeneralist2024}        & 47.2 & 9.7 & 4.2 & 56.9 & 30.0 \\
SpatialVLA\citep{quSpatialVLAExploringSpatial2025a}        & 16.7 & 25.0 & \textbf{29.2} & \textbf{100.0} & 42.7 \\
CogACT~\citep{liCogACTFoundationalVisionLanguageAction2024a}            & 58.3 & \underline{37.5} & \underline{20.8} & 91.7 & \underline{52.1} \\
Pi-0-Fast*\cite{pertschFastEfficientAction2025} & \underline{62.5} & 29.2 & \underline{20.8} & 83.3 & 49.0 \\
Pi-0.5*\citep{intelligence$p_05$VisionLanguageActionModel2025a}  & 58.3 & 33.3 & 16.7 & \underline{95.8} & 51.0 \\
AR-VLA (Ours)            & \textbf{75.0} & \textbf{54.2} & \underline{20.8} & \underline{95.8} & \textbf{61.5} \\
\bottomrule
\end{tabular}
\label{tab:simpler_vla}
\end{table}

\begin{table}[t]
\centering
\caption{Specialist Policy Performance. We report average max IOU and success rate (\%) for pushT, and success rate (\%) for two ALOHA task training on scripted and human demonstrations. Per seed evaluation results with standard error is accessible from our project website.}
\label{tab:specialist_performance}
\begin{tabular}{lcccccc}
\toprule
\multirow{2}{*}{\textbf{Method}} & \multicolumn{2}{c}{\textbf{pushT}} & \multicolumn{2}{c}{\textbf{aloha-cube}} & \multicolumn{2}{c}{\textbf{aloha-insert}} \\
\cmidrule(lr){2-3} \cmidrule(lr){4-5} \cmidrule(lr){6-7}
& Max IoU & Success & Script & Human & Script & Human \\
\midrule
DP\cite{chiDiffusionPolicyVisuomotor2023a}            & \textbf{0.957} & {\textbf{65.20}} & 33.33 & 10.00 & 22.67 & 1.66 \\
ACT\cite{zhaoLearningFinegrainedBimanual2023b} & 0.800          & 52.00       & \underline{86.00} & \underline{50.00} & \underline{32.00} & \textbf{20.00} \\

AR(Ours)                 & \underline{0.920} & \underline{60.40} & \textbf{97.33} & \textbf{67.33} & \textbf{54.67} & \underline{6.67} \\
\bottomrule
\end{tabular}
\end{table}

\begin{figure*}[ht]
    \centering
    \includegraphics[width=0.9\linewidth]{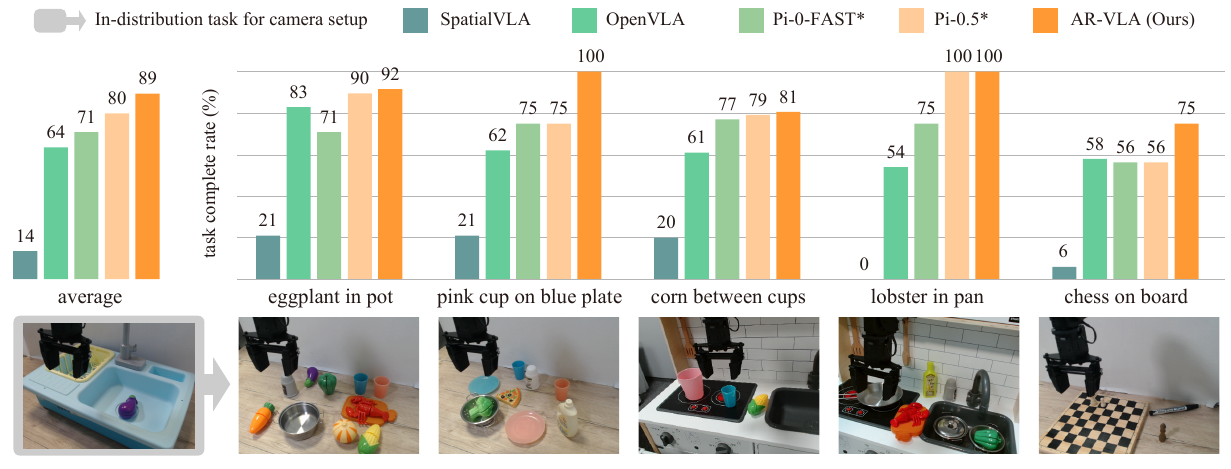}
    \caption{\textbf{BridgeV2 pretraining to real-world WidowX Zero-Shot Performance Comparison.} We set the camera pose so that all methods, except SpatialVLA\cite{quSpatialVLAExploringSpatial2025a}, reach a 100\% success rate on an easy in-distribution task, then test them zero-shot on challenging tasks. Details of experiment protocol in Appendix. } 
    \label{fig:widowx}
        \vspace{-10pt}
\end{figure*}

\begin{figure}[t]
    \centering
    \includegraphics[width=0.9\linewidth]{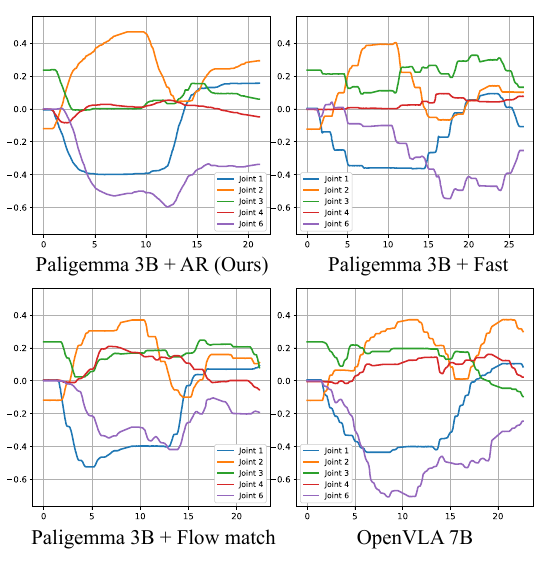}
    \caption{\textbf{Smoothness Visualization.} Joint states captured from success execution for the same task. X-axis in seconds, y-axis in rad}
    \label{fig:Joints}

\end{figure}
\smallskip
\noindent\textbf{Real-world WidowX Evaluation.}
We compare the zero-shot real-world performance of AR-VLA against state-of-the-art generalist policies, including OpenVLA and our reimplementation of the Pi family (Fig. \ref{fig:widowx}). Our Action Expert achieves a superior 89\% average success rate, notably reaching 100\% success on the ``cup on plate" and ``Lobster" tasks. A key driver of this performance is the model's inherent temporal awareness: upon an initial failure, AR-VLA gracefully lifts the end-effector to attempt a second trial. In contrast, other baselines often exhibit erratic motions near the target after a failed grasp, frequently pushing the object into configurations from which the policy cannot recover. This demonstrates that our autoregressive approach provides the closed-loop robustness necessary for reliable physical execution.

\smallskip
\noindent\textbf{Specialist Policy Performance.}
Table~\ref{tab:specialist_performance} shows AR Actor achieves strong performance across all tasks, demonstrating its viability as a robust policy backbone for task-specific imitation learning. On ALOHA cube transfer, AR Actor reaches 97.33\% scripted success and 67.33\% human demonstration success, substantially outperforming ACT (86.0\%/50.0\%) and Diffusion Policy (33.33\%/10.0\%). On ALOHA peg insertion, AR Actor achieves 54.67\% scripted success versus ACT's 32.0\%. On PushT, while Diffusion Policy achieves the highest success rate of 65.20\%, AR Actor reaches competitive 60.40\% with 0.920 Max IoU.
Notably, AR Actor maintains consistent performance across diverse task types, ranging from 2D pushing manipulation to bi-manual insertion tasks, while other methods show significant task-specific variance. Our AR policy is trained with a single objective: to maximize the conditional likelihood of each action in a sequence. This suggests that architectural simplicity may be a key factor for consistent task-agnostic performance, whereas more complex methods like Diffusion Policy excel on specific tasks but struggle on others. Our observations align with similar results that have been reported by ARP \cite{zhang2025autoregressiveactionsequencelearning}.

\begin{table}[t]
\centering
\caption{\textbf{Inference Smoothness Metrics.} Jerk in $10^2$ rad/s$^3$. VLM and Action Expert latencies in  ms / chunk size. Total/Act represents the effective latency per single action.}
\setlength{\tabcolsep}{4pt}
\begin{tabular}{lccccc}
\toprule
\textbf{Model} & \multicolumn{2}{c}{\textbf{Jerk ($\downarrow$)}} & \multicolumn{3}{c}{\textbf{Latency ($\downarrow$)}} \\
\cmidrule(lr){2-3} \cmidrule(lr){4-6}
& \text{Avg} & \text{Max} & \text{VLM} & \text{Act. Expert} & \text{Total/Act} \\
\midrule
OpenVLA   & 10.13            & \underline{42.14}            & \multicolumn{2}{c}{321.72 / 1}       & 321.72          \\
Fast      & \underline{8.15} & 80.24            & \multicolumn{2}{c}{744.78 / 4}       & 186.20          \\
FM        & 9.39             & 45.33 & 69.77 / 4 & \underline{267.28 / 4} & \underline{84.26} \\
AR (Ours) & \textbf{7.89}    & \textbf{39.83}    & 69.56 / 4 & \textbf{28.86 / 1} & \textbf{46.25}  \\
\bottomrule
\end{tabular}
\vspace{-15pt}
\label{tab:inference_with_total}
\end{table}

\begin{figure*}[t]
    \centering
    \includegraphics[width=1.0 \linewidth]{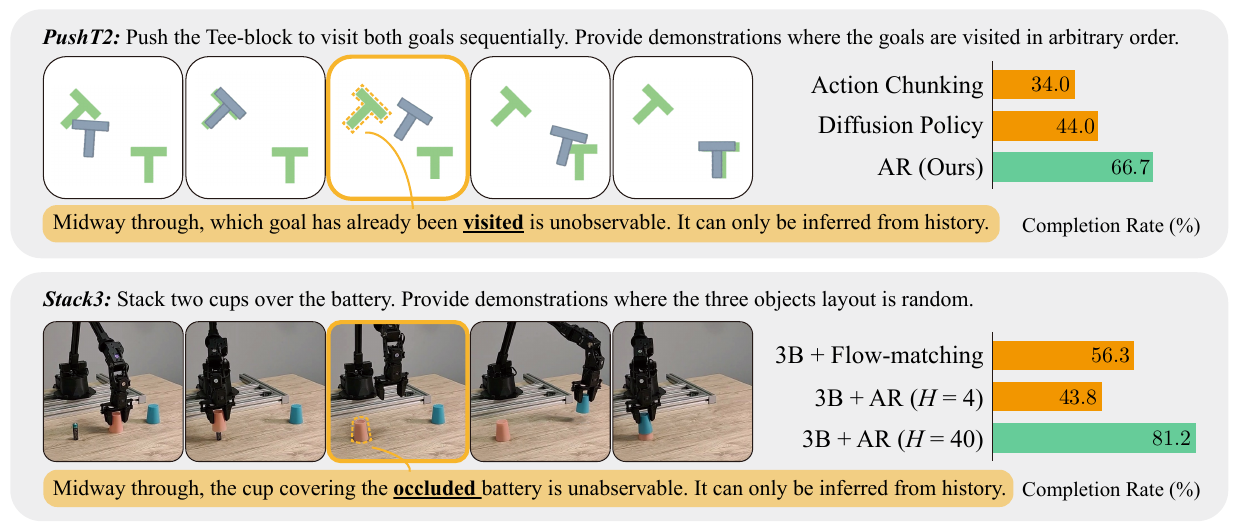}
    \caption{\textbf{History-Awareness Evaluation.} PushT2 requires visiting both goals, but which goal has been visited is unobservable midway. Stack3 requires stacking cups over a battery that becomes occluded. Both task require memory of unobservable past states. $H$ donates the context window length of AR-VLA. Details about task defination, data collection, training and execution in Appendix. }
    \label{fig:history_tasks}
    \vspace{-15pt}
\end{figure*}

\subsection{Efficiency and Smoothness Analysis} \label{sec:smoothness}
We evaluate the systemic advantages of our asynchronous execution framework. By structurally decoupling the control thread from VLM inference, AR-VLA maintains a stable 29ms per action control frequency even when the perception backbone is capped at 70ms per frame. As shown qualitatively in and Fig. \ref{fig:Joints} quantitatively in Tab.~\ref{tab:inference_with_total}, AR-VLA achieves the lowest maximum and average jerk during task execution. This superiority arises from the model's ability to maintain rigorous spatial and temporal action consistency, alongside the lowest overall latency. In contrast, chunk-based models, when implemented naively, only guarantee intra-chunk smoothness and suffer from inter-chunk gaps and execution latency.

\subsection{History-Awareness Evaluation} \label{sec:history}
To validate that autoregressive action generation with stored history key-value caches enables temporal awareness for long-horizon tasks, we design two benchmark tasks requiring memory of past states. As shown in Fig.~\ref{fig:history_tasks}, \textit{PushT2} (simulation) requires pushing a T-shaped block to two goal positions in any order, where information about which goal has been reached becomes unobservable once the block is halfway to the second goal. \textit{Stack3} (real-world) requires covering a battery with one cup then stacking another cup on top, where the battery's location becomes unobservable once covered and can only be retrieved from action history.
AR-VLA substantially outperforms existing policies on both tasks. Reactive policies exhibit ``temporal amnesia," becoming trapped oscillating between sub-goals without maintaining context of completed steps. In contrast, AR-VLA's autoregressive generation naturally maintains history, momentum, and task intent through its action domain key-value cache, leading to superior success rates on these multi-stage manipulation sequences.

\subsection{Ablations on Design Decisions}
In this section, we conduct ablation studies to validate the contributions of our key design choices. Table~\ref{tab:ablation_compact} presents the results across four critical components.

\noindent \textbf{Causal Pretraining.} We compare models with and without Phase 1 pretraining. Removing Phase 1 pretraining increases Phase 2 convergence time to 2$\times$ and still leads to inferior final performance, as the model must learn joint kinematics from scratch rather than building upon pretrained motion priors.

\noindent \textbf{RoPE Re-anchoring.} We ablate different positional encoding strategies for the action expert. Replacing our dynamic RoPE anchoring with fixed rotational embedding or non-rotational embedding is mathematically incorrect and results in poor inference performance. While eliminating positional embedding avoids training-inference attention score gaps, it cannot capture temporal step differences, resulting in a 52\% performance drop. This validates that {proper temporal position encoding is critical for autoregressive action generation}.

\noindent \textbf{Stochastic History Masking.} We observe an intriguing causal confusion paradox across different masking rates. A $0.0$ mask rate achieves the lowest validation error but $0\%$ task success, indicating the model over-relies on its own history and fails during rollouts when predictions deviate. Our $0.6$ mask rate provides the optimal balance between leveraging historical context and maintaining robustness to prediction errors.

\noindent \textbf{Context Length.} We vary the action KV cache length from $1$ to $40$ steps. The results show a clear upward trend in success rate as context length increases, directly validating that {temporal awareness through extended history correlates with improved performance for autoregressive action experts}.

\begin{table}[t]
\centering
\caption{Ablation Study of AR-VLA Design Choice. Validate Error calculate from the model output and ground truth when taking dataset trajectory as inout. Success rates are percentages in total 96 SIMPLER trials after training. Bold indicates the full AR-VLA configuration.}
\label{tab:ablation_compact}
\begin{tabular}{lcc}
\toprule
\textbf{Variant} & \textbf{Val. Error ($\downarrow$)} & \textbf{Sim. Success ($\uparrow$)} \\ 
\midrule
w/ Phase-1 \textbf{(Full)} & 4.2 & 61.5 \\
w/o Phase-1 (100\% time) & 4.5 & 37.5 \\
w/o Phase-1 (200\% time) & 4.0 & 54.2 \\ 
\midrule
mask rate = 0.0 & 2.7 & 0.0 \\ 
mask rate = 0.2 & 3.1 & 27.1 \\ 
mask rate = 0.4 & 3.5 & 47.9 \\ 
mask rate = 0.6 (\textbf{Full}) & 4.2 & 61.5 \\ 
mask rate = 0.8 & 5.7 & 49.0 \\ 
mask rate = 1.0 & 2.9 & 28.1 \\ 
\midrule
w/ Static Pos. Embedding & 3.0 & 3.1 \\ 
w/o Pos. Embedding & 2.9 & 29.2 \\ 
\midrule
history length = 1 & 6.0 & 36.5 \\ 
history length = 5 & 5.2 & 50.0 \\ 
history length = 10 & 4.7 & 59.4 \\ 
history length = 20 (\textbf{Full}) & 4.2 & 61.5 \\ 
history length = 40 & 4.2 & 59.4 \\ 
\bottomrule
\end{tabular}
\vspace{-10pt}
\end{table}

\section{Conclusion and Discussion} 
\label{sec:conclusion}
We introduced a design of Autoregressive Action Expert, facilitating a structural shift from the prevailing paradigm of reactive, snapshot-based control to continuous streaming sequences. By equipping the policy with a persistent causal history of its own, our approach resolves the temporal inconsistencies inherent in current VLA architectures and ensures action generation remains consistent across both spatial and temporal dimensions.
Empirically, we demonstrate that this autoregressive structure yields significantly smoother trajectories and reduced execution latency compared to reactive baselines, with particular robustness in long-horizon tasks where maintaining historical context is critical. Beyond immediate performance gains, our architecture offers a scalable framework for embodied learning by structurally decoupling the syntax of motion from semantic perception. This allows for the independent pretraining of kinematic dynamics and the asynchronous integration of perception and control. 

However, transitioning to this autoregressive, cache-based paradigm introduces new challenges. First, novel sequences of individually familiar states within the KV-cache can compound, pushing the policy into out-of-distribution (OOD) failures. Second, similar to flow-matching experts, direct AR action gradients degrade the VLM's pretrained semantic priors, necessitating a ``knowledge insulation'' \cite{intelligence$p_05$VisionLanguageActionModel2025a} training strategy. Finally, while proprioceptive history is fully cached, visual processing remains restricted to standard snapshots. We detail these limitations and future directions in Appendix.

\section*{Acknowledgments}
This research was partially funded by the Ministry of Education and Science of Bulgaria (support for INSAIT, part of the Bulgarian National Roadmap for Research Infrastructure), and supported by Interne Fondsen KU Leuven / Internal Funds KU Leuven (C2E/24/034).

\bibliographystyle{plainnat}
\bibliography{ARVLA-RSS2026}

\newpage
\clearpage
\onecolumn 

\begin{adjustwidth}{0.4in}{0.4in}
\captionsetup{width=6.5in, justification=centering}

\linespread{1.2}\selectfont
\fontsize{11pt}{13pt}\selectfont


\makeatletter
\renewcommand{\section}{\@startsection{section}{1}{\z@}%
    {-3.5ex \@plus -1ex \@minus -.2ex}%
    {2.3ex \@plus.2ex}%
    {\normalfont\Large\bfseries}} 

\renewcommand{\subsection}{\@startsection{subsection}{2}{\z@}%
    {-3.25ex\@plus -1ex \@minus -.2ex}%
    {1.5ex \@plus .2ex}%
    {\normalfont\large\bfseries}} 
\makeatother

\section*{\Large Appendix}
\label{sec:appendix_main}
Please check the supplementary material for videos for different task, and an animation of how the model works. We provide below additional tenchical details about:

\begin{itemize}
    \item \textbf{Section~\ref{sec:app_arch}: Model Architecture} -- Detailed layer configurations for PaliGemma, the Action Expert, strategy to guarantee fair comparison against other generalist and specialist policies.
    \item \textbf{Section~\ref{sec:app_hyperparams}: Hyperparameters} -- Complete config tables for Generalist and Specialist models (Learning rates, batch sizes, and sequence lengths).
    \item \textbf{Section~\ref{sec:app_tasks}: Benchmark Protocol} -- Explanation of the scoring rubric and the "eggplant-in-sink" calibration for real-world WidowX experiments.
    \item \textbf{Section~\ref{sec:app_history_task}: History Awareness Tasks} -- Design and data collection details for the \textit{PushT2} and \textit{Stack3} non-Markovian benchmarks.
    \item \textbf{Section~\ref{sec:app_screenshots}: Execution Highlights} -- Visual snapshots of model performance across simulation and real-world experiments.
    \item \textbf{Section~\ref{sec:app_limitations}: Discussion \& Limitations} -- An analysis of compounding errors, gradient dynamics, and the future of streaming VLMs / VLAs.
\end{itemize}

\subsection{Model Architecture}
\label{sec:app_arch}

The VLA system is built upon the PaliGemma vision-language framework, which integrates a SigLIP-So400m visual encoder with a Gemma-2b language model. The visual tower consists of 27 transformer encoder layers utilizing 14x14 patch embeddings with a hidden dimension of 1152 and a feed-forward dimension of 4304. The language model component is an 18-layer causal transformer decoder with a hidden dimension of 2048 and a feed-forward dimension of 16,384. To maintain the semantic reasoning capabilities acquired during large-scale pretraining, both the vision tower and the language model are kept frozen throughout the learning process.

The core of our method is the auto-regressive Action Expert, this is achieved by redefine the prediction target and runtime behavior, on the exact same structure which specifically designed by Pi-zero to parallel the structural depth of the language backbone. The expert consists of 18 transformer layers, intentionally matching the layer count of the Gemma-2b component. This symmetry facilitates the alignment of visual-linguistic semantics with kinematic sequences, allowing the expert to attend to the vision-language KV cache across corresponding layers. Despite this structural parity, the Action Expert is significantly more parameter-efficient, utilizing a reduced hidden dimension of 1024 and an MLP dimension of 4096. The attention mechanism within the expert employs 2048-dimensional query projections and 256-dimensional key/value projections, integrated with Rotary Positional Embeddings (RoPE) to ensure precise relative temporal anchoring. This decoupled design allows the high-frequency Action Expert to generate control streams while conditioning on the lower-frequency semantic signals provided by the PaliGemma backbone.

\begin{figure}[h]
    \centering
    \includegraphics[width=0.95\linewidth]{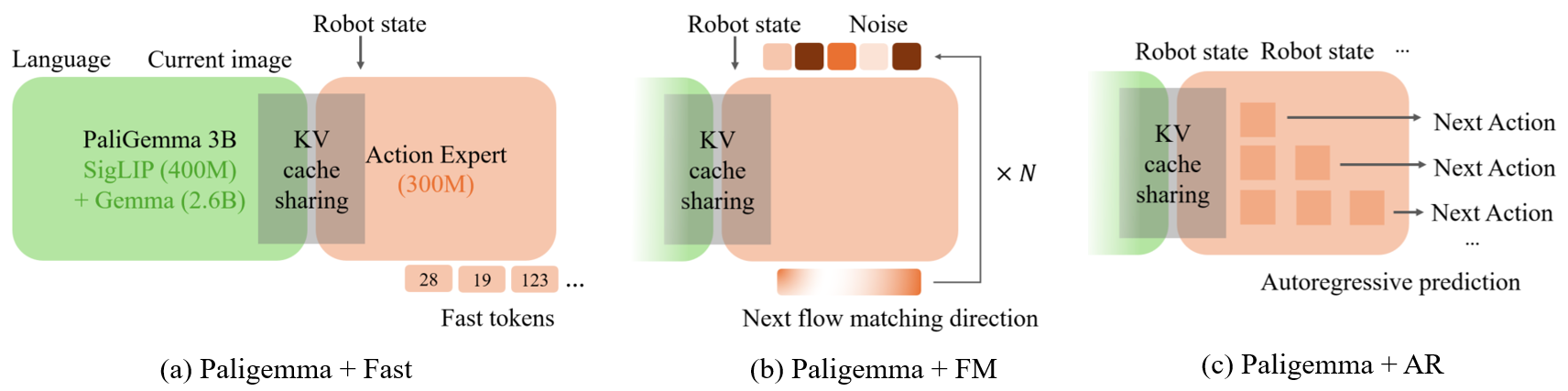}
    \caption{Three different Action Experts sharing the same architecture and V-L backbone. The same networks are trained and used differently.}
\end{figure}

To evaluate the effectiveness of the proposed autoregressive streaming approach, we implemented a baseline following the $\pi_{0.5}$ training recipe adapted for the BridgeV2 dataset. Both the baseline and our proposed AR-VLA maintain an identical model scale, consisting of a PaliGemma-3B vision-language backbone and a 300M-parameter transformer-based Action Expert.

We strictly adhered to a knowledge insulation strategy to ensure that semantic reasoning and motor control are learned through independent signals. Specifically, the VLM parameters were trained exclusively using an auxiliary Fast Token loss. Gradients originating from the action generation objective were blocked from entering the VLM layers, ensuring that the action loss only updated the parameters of the Action Expert itself.

For the AR-VLA Action Expert, we utilized the exact architectural configuration and parameter budget as the transformer model employed for flow matching in the $\pi_{0.5}$ baseline. The primary distinction lies in the generative objective; while the baseline utilizes multi-round iterative denoising (flow matching) to generate action chunks, our Action Expert employs a standard next-action prediction loss to generate actions autoregressively in a single inference step. This methodology allows for a controlled comparison between three models of the same 3.3B scale: $\pi_{0\text{-FAST}}$, which generates actions by decoding Fast tokens; $\pi_{0.5}$, which predicts action chunks through iterative flow matching; and AR-VLA, which generates actions sequentially as a continuous autoregressive stream. Note here the Pi-0-FAST model do not strictly follow the original design --- we use a separate 300M Action Expert --- instead of the VLM itself to generate fast tokens. This is because (1) The gemma architecture is too heavy and takes too much time to generate the whole fast token piece for a chunk (2) We want to compare the different action expert formulation using the same parameter scale.

\begin{figure}[h]
    \centering
    \includegraphics[width=0.95\linewidth]{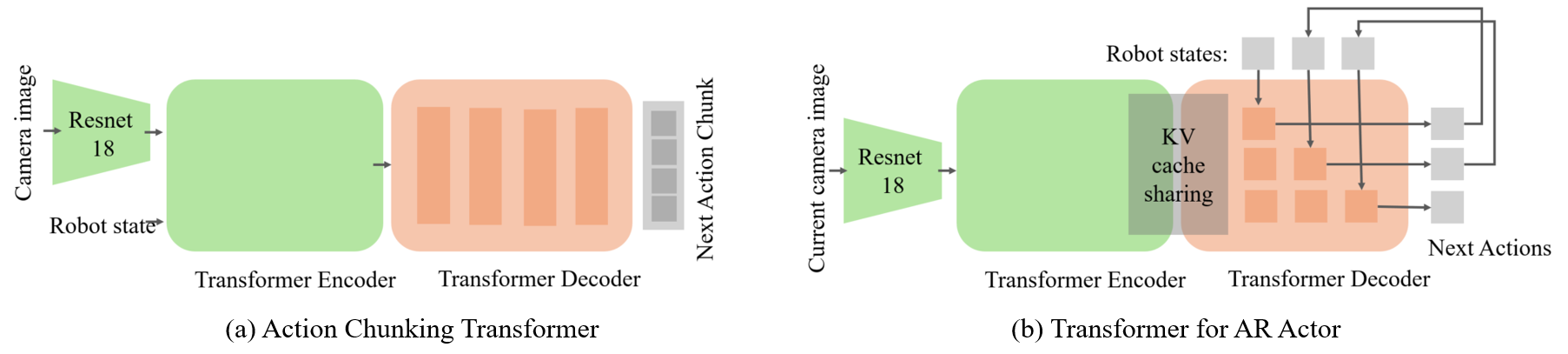}
    \caption{AR Actor that share the exact same size and architecture of Action Chunking Transformer, the same decoders are trained and used differently.}
    \label{fig:placeholder}
\end{figure}

For our specialist policy evaluations, we compared our approach against two state-of-the-art specialist frameworks, namely Action Chunking Transformer (ACT) and Diffusion Policy. In this setting, we utilized the LeRobot implementations for both baselines, where all methods shared a common ResNet-18 vision encoder, and we implement our AR actor in a structure comparable to the ACT implementation.

The adaptation of the ACT architecture into our autoregressive framework was conducted while maintaining strict parity in parameter size and structural configuration. While the original ACT utilizes a transformer-based configuration consisting of a 4-layer encoder and a 4-layer decoder to predict action chunks, we modified this design to function as an AR-VLA-like system. The 4-layer encoder was treated as a localized perception backbone, and its intermediate key-value pairs were cached to provide a semantic prefix. The 4-layer decoder was subsequently repurposed as a standalone autoregressive Action Expert. By replacing the original chunk-based objective with a standard next-token prediction loss, we enabled the model to perform streaming control while leveraging the same underlying transformer architecture as the specialist baseline.

\subsection{Hyperparameters}
\label{sec:app_hyperparams}

The generalist AR-VLA model is built upon a FAST-insulated PaliGemma-3B backbone, while the specialist models utilize a ResNet-18 vision backbone. In both cases, the Action Expert is designed to cross-attend to the perception features while maintaining a persistent internal history.

\begin{table}[htbp]
\centering
\caption{Architecture specifications for Generalist (AR-VLA) and Specialist (AR-Actor) models. The specialist configuration is based on the 4-4 Transformer-based Actor setup.}
\label{tab:arch_comparison}
\begin{tabular}{lll}
\toprule
\textbf{Parameter} & \textbf{Generalist (AR-VLA)} & \textbf{Specialist (AR-Actor)} \\ 
\midrule
Vision Backbone & SigLIP-So400m & ResNet-18 \\
Language/State Backbone & Gemma-2b (Fast) & Transformer Encoder (Trainable) \\
Action Expert Layers & 18 Transformer Decoder & 4 Transformer Decoder \\
Hidden Dimension ($D$) & 1024 & 512 \\
Feed-forward Dim ($D_{ff}$) & 4096 & 3200 \\
Attention Heads & 8 & 8 \\
Positional Encoding & RoPE (Dynamic) & RoPE (Dynamic) \\
Activation Function & GELU (tanh) & ReLU \\
Dropout Rate & 0.0 & 0.1 \\
Action Space Dim & 8 (3trans + 4quaterion + 1gripper) & 14 (Joint Space) \\
\bottomrule
\end{tabular}
\end{table}

Training for generalist models follows a two-phase strategy. Phase-1 focuses on mastering the kinematic syntax using action-only data, while Phase-2 performs multimodal alignment. Phase-1 is notably efficient, requiring only 20,000 steps with a batch size of 1024; this stage typically completes within approximately 2 hours on a single A6000 GPU. Crucially, the action-only dataset used for Phase-1 is curated to be distinct from the BridgeV2 dataset used in the final alignment. Specialist models are trained using a standard offline imitation learning pipeline.

\begin{table}[htbp]
\centering
\caption{Comparative training configurations for generalist (Phases 1 \& 2) and specialist policies.}
\label{tab:training_configs}
\begin{tabular}{llll}
\toprule
\textbf{Hyperparameter} & \textbf{VLA (Phase 1)} & \textbf{VLA (Phase 2)} & \textbf{Specialist} \\ 
\midrule
Optimizer & AdamW & AdamW & AdamW \\
Learning Rate (LR) & $1.0 \times 10^{-4}$ & $5.0 \times 10^{-5}$ & $1.0 \times 10^{-5}$ \\
LR Backbone & - (N/A) & - (Insulated) & $1.0 \times 10^{-5}$ \\
Weight Decay & 0.0 & 0.0 & $1.0 \times 10^{-4}$ \\
Grad. Clip Norm & 1.0 & 1.0 & 10.0 \\
Batch Size & 1024 & 512 & 8 \\
Total Training Steps & 20,000 & 30,000 & 200,000 \\
Warmup Steps & 500 & 200 & 500 \\
Precision & BFloat16 & BFloat16 & Float32 \\
Normalization & Mixed (Q99) & Mixed (Q99) & Mean/Std \\
\bottomrule
\end{tabular}
\end{table}
To maintain temporal consistency, we employ specific history and masking parameters. The generalist model is optimized for long-horizon stability with a larger history window, while the specialist model operates at higher frequencies for precise bimanual tasks.

\begin{table}[htbp]
\centering
\caption{Sequence and execution parameters across environments.}
\label{tab:inference_params}
\begin{tabular}{lll}
\toprule
\textbf{Parameter} & \textbf{BridgeV2 (Generalist)} & \textbf{ALOHA Sim (Specialist)} \\ 
\midrule
Training History Length ($H$) & 16 Steps & 20 Steps \\
Causal Training Length ($H$) & 8 Steps & 20 Steps \\
Test-Time History & 20 Steps & 30 Steps \\
History Masking Prob. & 0.6 & 0.5 \\
\bottomrule
\end{tabular}
\end{table}

The loss weights for AR-VLA are distributed as: Translation ($\lambda=1.0$), Rotation ($\lambda=1.0$), Gripper ($\lambda=0.1$). For the specialist models, we optimize a standard L2 loss on next-action prediction with the exclusion of VAE-based latent modeling to focus purely on the autoregressive sequence dependency.

\subsection{Benchmark Protocol and Completion Criteria}
\label{sec:app_tasks}

Our evaluation protocol is designed to provide a rigorous assessment of model performance across both simulated and real-world environments. For each task, we define $M$ distinct initial conditions by varying the starting positions of relevant objects and the robot's end-effector. We execute $N$ trials for each initial condition, resulting in a total of $N \times M$ trials per task. Performance is reported as the average task progress across all tasks, configurations, and trials. Following established benchmarks in the field, we utilize a scoring rubric that assigns partial credit (0.25, 0.50, 0.75, or 1.00) based on the successful completion of specific task milestones, as detailed in Table~\ref{tab:scoring_rubric}.

For the real-world WidowX experiments, our hardware setup closely mirrors the BridgeV2 environment. We utilize a single external camera positioned to the side of the robot arm, directed toward the workspace. While VLA models generally exhibit robustness to camera pose variations, we implement a calibration step to ensure a fair comparison: before benchmarking, we verify that each model achieves a 3/3 success rate on an ``eggplant in sink'' task using the exact toy kitchen setup from the training dataset. This ensures the chosen camera pose is viable for all competing architectures. 

Real-world episodes are subject to a timeout of 200 steps (equivalent to 40 seconds at a 5Hz control frequency). Episodes are only manually terminated if the robot enters a state that risks hardware damage; otherwise, collisions or unintended object displacements are permitted. The final score for an episode is determined by the highest completion state reached during the trial according to the scoring criteria.

\begin{table}[htbp]
\centering
\caption{Scoring rubric for real-world WidowX evaluation tasks. Scores represent cumulative progress toward task completion.}
\label{tab:scoring_rubric}
\begin{tabular}{lcccc}
\toprule
\textbf{Task} & \textbf{0.25} & \textbf{0.50} & \textbf{0.75} & \textbf{1.00} \\ 
\midrule
Eggplant in Pot & Reach eggplant & Pick up eggplant & Drop near pot & Place in/on pot \\
Pink Cup on Plate & Reach pink cup & Pick up pink cup & Drop near plate & Place on plate \\
Chess Piece on Board & Reach piece & Pick up piece & - & Place on board \\
Lobster in Pan & Reach lobster & Pick up lobster & Drop near pan & Place in pan \\
Corn Between Cups & Reach corn & Pick up corn & - & Place between cups \\
\bottomrule
\end{tabular}
\end{table}

\subsection{History Awareness Task: Design, Data Collection, and Execution}
\label{sec:app_history_task}

We designed two specific benchmark tasks requiring memory of non-observable past states: \textit{PushT2} in simulation and \textit{Stack3} in the real world. These tasks are spatially configured such that success depends on retrieving information from the action history rather than immediate visual feedback.

The \textit{PushT2} task (Simulation) requires the robot to push a T-shaped block to two distinct goal positions in any order. The information regarding which goal has already been occupied becomes unobservable once the block moves halfway toward the second goal. The \textit{Stack3} task (Real-world) requires the robot to cover a battery with one cup and subsequently stack a second cup on top of the first. Once the battery is covered by the initial cup, its location is no longer visible in the camera stream; the robot must rely on its internal history to identify which cup contains the battery and where to place the second cup.

\begin{figure}[h]
    \centering
    \includegraphics[width=0.9\linewidth]{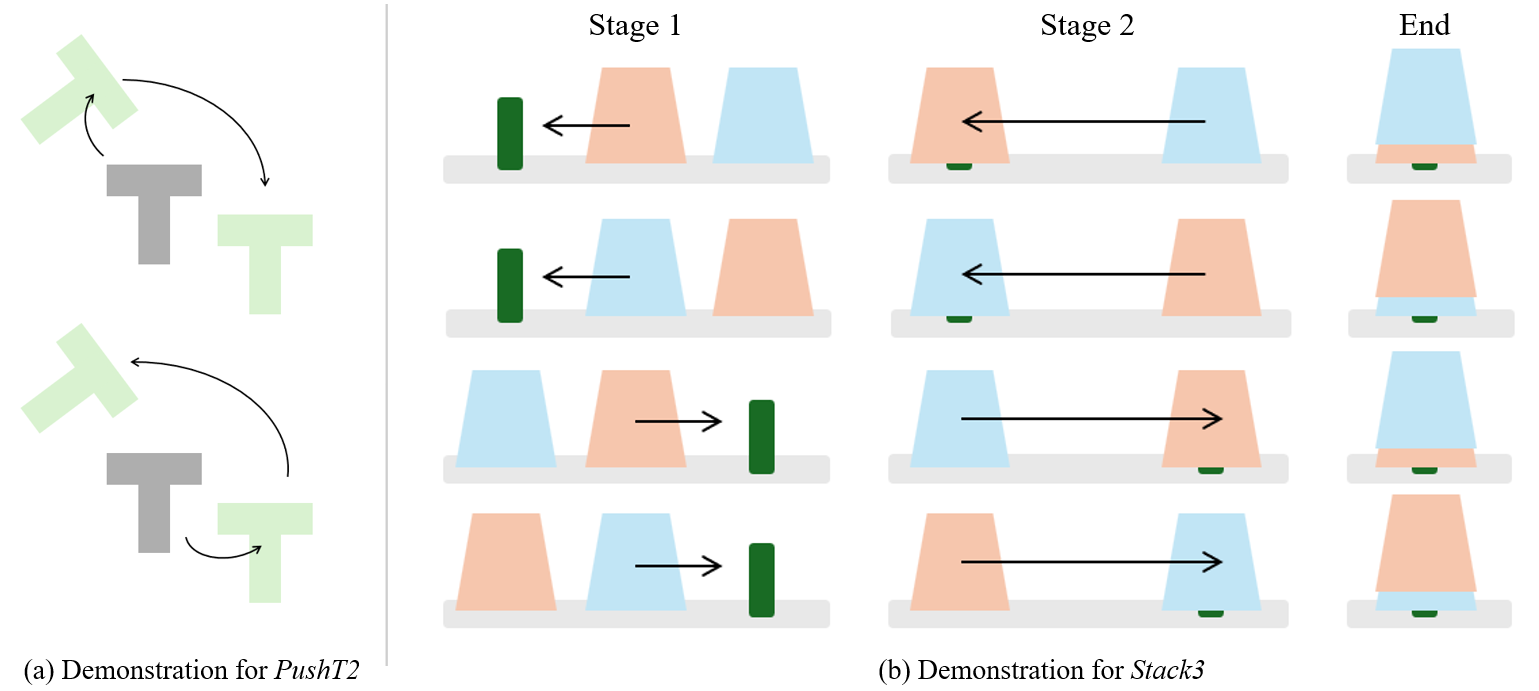}
    \caption{Demonstration collection for history-aware tasks. (a) The \textit{PushT2} task can be complete in two different orders, we don't constrain human demonstrators on their solution. (b) We collect demonstrations for \textit{Stack3} under different initial object layouts. On stage 2, the visual snapshot is identical (the battery in dark is covered, and the camera can only see two cups) but the next object to pick is different.}
\end{figure}

For the \textit{PushT2} environment, we collected 240 episodes from human experts. These demonstrations were performed without a mandated target-reaching order, allowing the model to learn a flexible multi-goal intent. For the real-world \textit{Stack3} task, we collected a total of 16 teleoperated episodes. These demonstrations utilized varying layout orders for the two cups and the battery to ensure the policy generalizes across different initial configurations while maintaining the hidden midway state.

We define rigorous completion metrics for both tasks to quantify the benefits of history maintenance:

\begin{itemize}
    \item \textbf{PushT2 (Simulation):} A target goal is considered successfully reached if the Intersection over Union (IoU) between the T-block and the goal area is greater than 0.9.
    \item \textbf{Stack3 (Real-world):} We utilize a tiered scoring rubric to evaluate multi-stage progress:
    \begin{itemize}
        \item \textbf{0.25:} Robot successfully picks up the first cup.
        \item \textbf{0.50:} Robot places the first cup over the battery (covering it).
        \item \textbf{0.75:} Robot identifies and picks up the second (correct) cup.
        \item \textbf{1.00:} Robot successfully stacks the second cup on top of the first.
    \end{itemize}
\end{itemize}

Experimental results indicate that reactive policies frequently fail these benchmarks due to ``temporal amnesia,'' resulting in oscillations between sub-goals or failure to locate hidden objects. In contrast, the AR-VLA architecture leverages its autoregressive generation and persistent KV cache to maintain task intent and kinematic momentum, resulting in significantly higher success rates on these non-Markovian manipulation sequences.

\begin{figure}[H]
    \centering
    \includegraphics[width=0.8\linewidth]{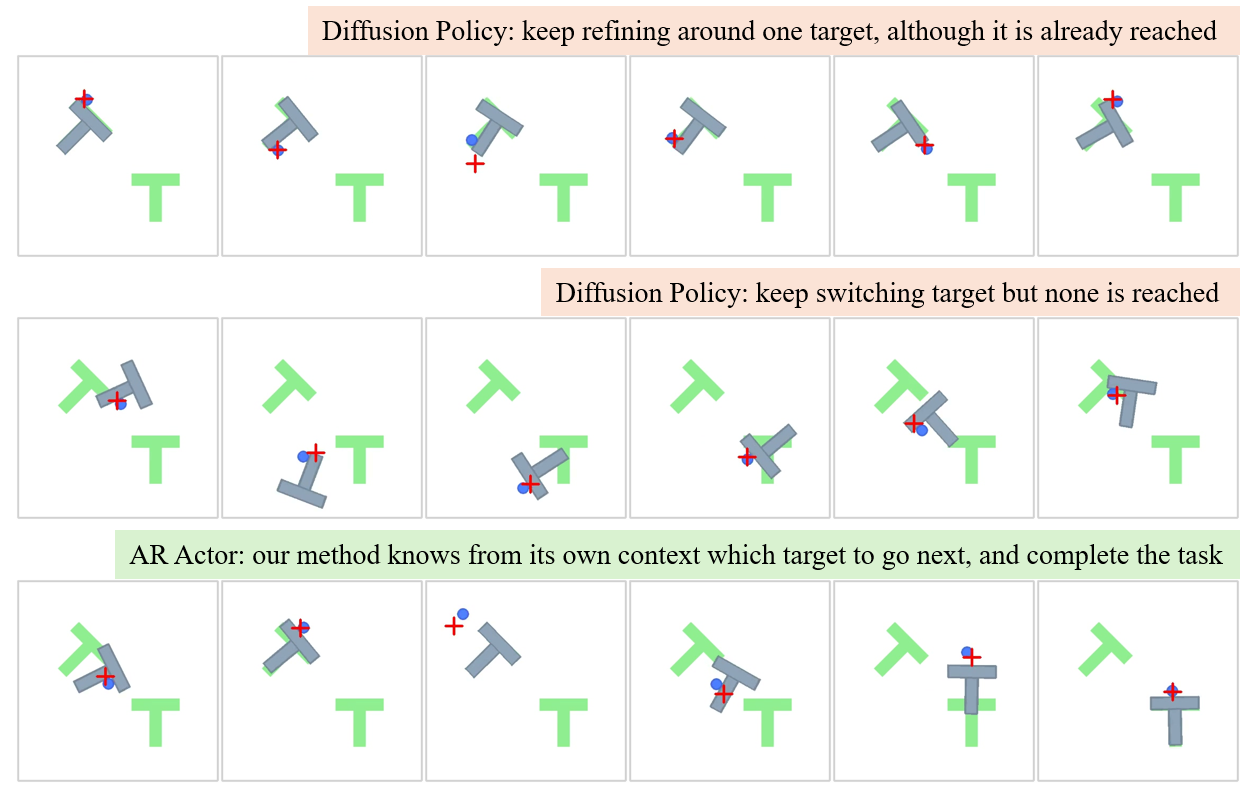}
    \caption{Typical cases during PushT2 task execution.}
\end{figure}

\begin{figure}[H]
    \centering
    \includegraphics[width=0.9\linewidth]{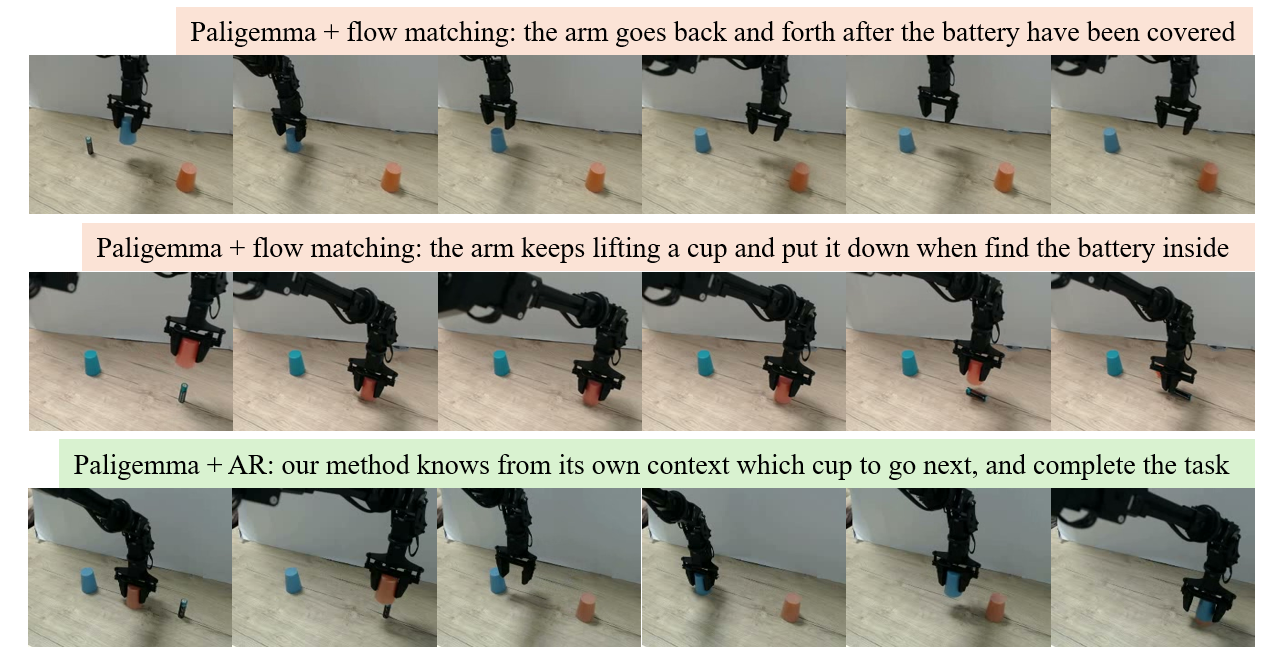}
    \caption{Typical cases during Stack3 task execution.}
\end{figure}

\subsection{Screenshot Highlights from Task Execution}
\label{sec:app_screenshots}

\begin{figure}[H]
    \centering
    \includegraphics[width=0.8\linewidth]{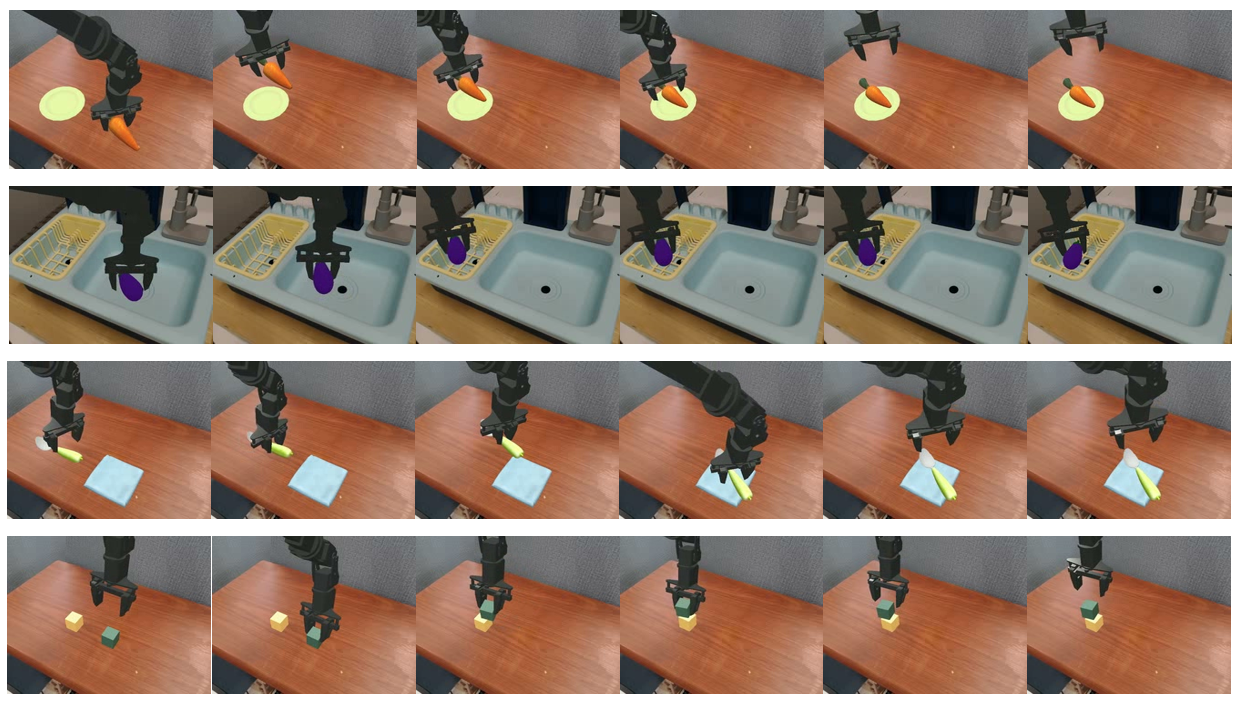}
    \caption{AR-VLA Zero-shot task execution in SIMPLER simulator.}
\end{figure}

\begin{figure}[H]
    \centering
    \includegraphics[width=0.8\linewidth]{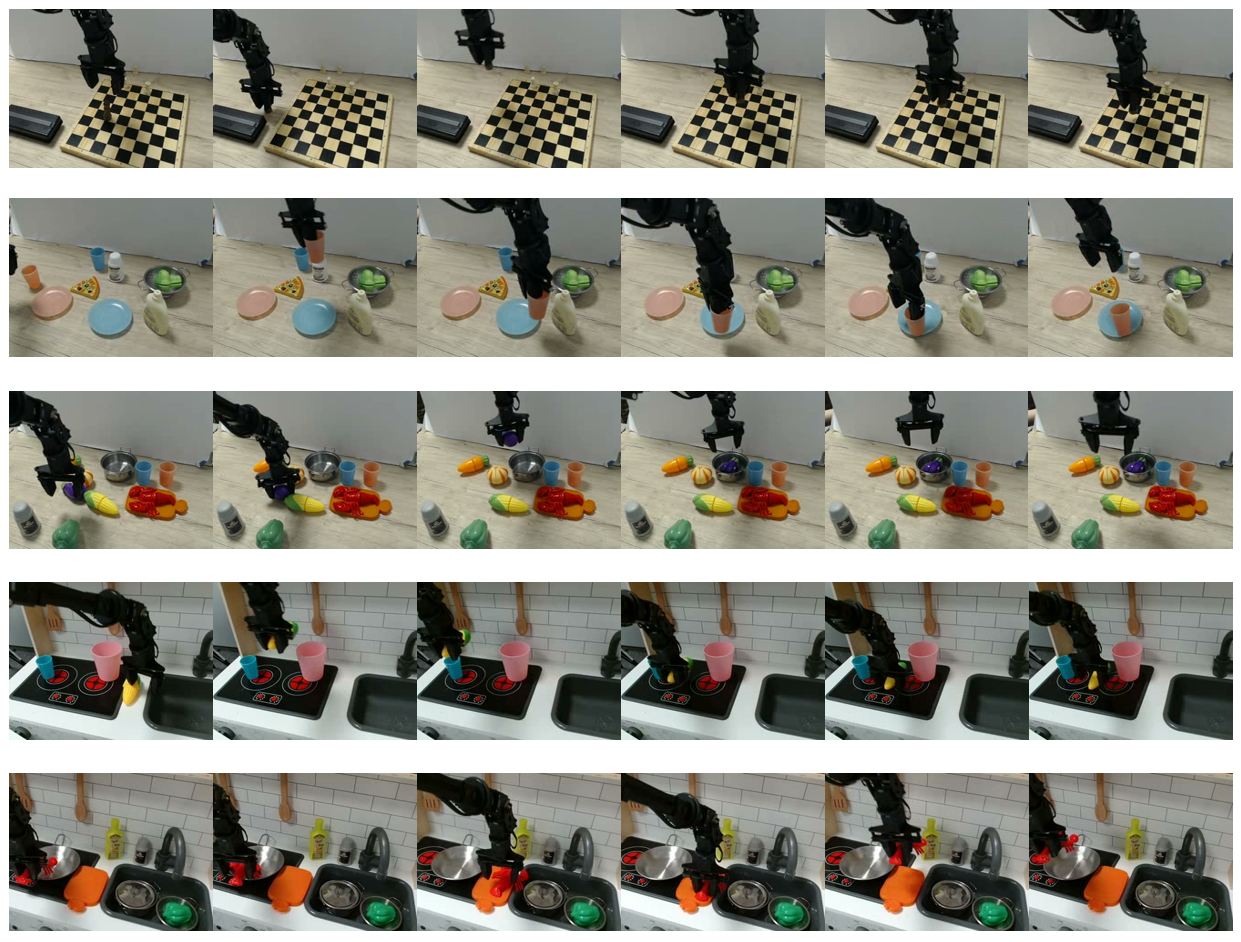}
    \caption{AR-VLA Zero-shot task execution in real world.}
\end{figure}

\begin{figure}[H]
    \centering
    \includegraphics[width=0.8\linewidth]{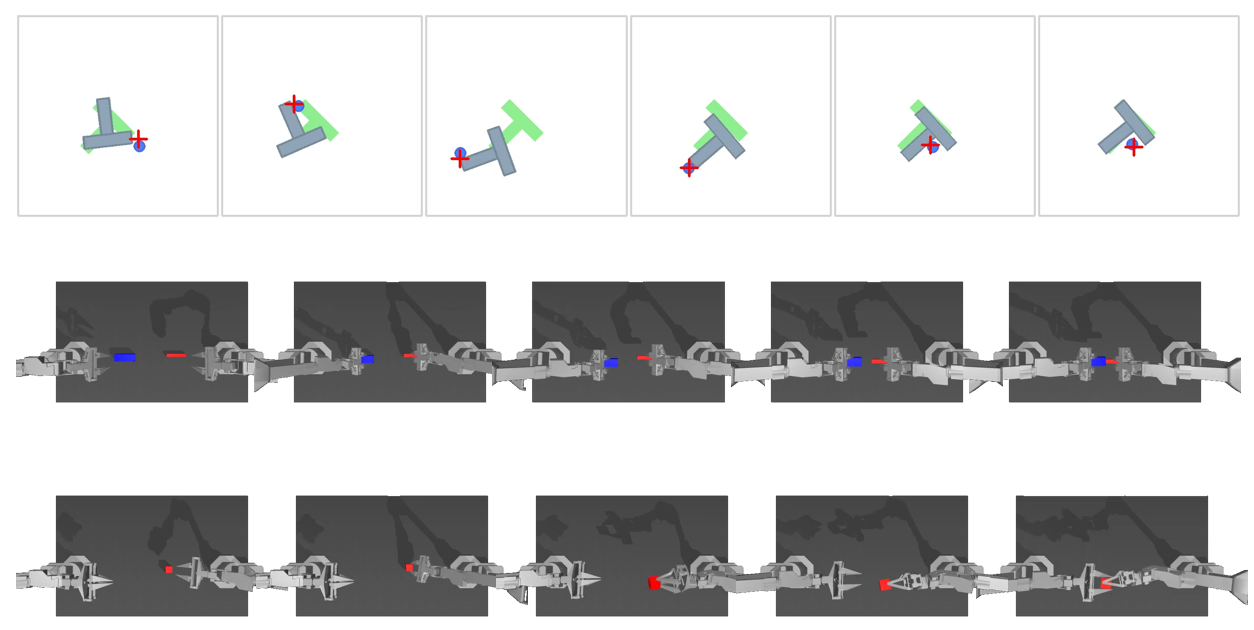}
    \caption{AR-Actor specialist task execution.}
\end{figure}

\subsection{Discussion and Limitations}
\label{sec:app_limitations}

While AR-VLA demonstrates significant improvements in temporal consistency and history awareness, several challenges and avenues for future research remain.

\textbf{Compounding Errors and OOD States.} 
A primary limitation of the autoregressive paradigm in robotics is its sensitivity to Out-of-Distribution (OOD) trajectories. Unlike reactive policies that ``reset" at every frame, AR models are functionally dependent on their own past predictions (That is partially why we need such a high history mask rate). If a model generates a slightly OOD action, that error is immediately encoded into the Key-Value cache as part of the kinematic history. This can trigger a deleterious feedback loop: the OOD history makes subsequent actions more unpredictable, which in turn pushes the robot further into OOD states. While our use of stochastic history masking during training partially mitigates this ``causal confusion", future work should investigate more robust recovery behaviors or the integration of uncertainty-aware sampling to break these failure loops.

\textbf{Gradient Dynamics and Knowledge Insulation.}
Our experiments confirm a phenomenon recently observed in flow-matching VLAs: gradients from the action-generation objective do not necessarily improve the semantic capabilities of the vision-language backbone. In fact, directly propagating autoregressive gradients into the VLM can be counter-productive, potentially degrading the rich semantic priors learned during large-scale pretraining. This justifies our ``knowledge insulation" strategy (borrowed from pi-0.5), where the VLM is kept frozen or updated via an independent auxiliary loss. Effectively pairing an AR actor with a fast-insulated VLM ensures that motor control and semantic reasoning do not compete for the same parameter updates, but it also suggests that finding a truly synergistic joint-training objective remains an open question.

\textbf{Integrated vs. Modular Autoregression.}
In this work, we instantiate the AR actor as a separate Action Expert. However, modern VLMs like PaliGemma are inherently suitable autoregressive in their language components. In principle, the LLM portion of a VLM could be used directly to model action sequences, similar to how it generates text. While our modular approach allows for higher control frequencies and independent pretraining, a fully integrated system where actions and language tokens are interleaved in the same causal stream represents a compelling alternative. Such a design could potentially lead to tighter coupling between semantic intent and physical execution.

\textbf{Streaming Not only VLA, but also VLM.}
Finally, the architectural principles of AR-VLA can be extended to the Vision-Language Model itself. Current VLM-based video modeling typically relies on a fixed number of screenshots processed in batches, which limits temporal resolution and real-time adaptability. Future VLMs could adopt our streaming approach to move beyond these discrete snapshots. In a "Streaming VLM" paradigm, the model could continuously refresh vision tokens in its KV cache in real-time as new frames arrive. Crucially, this would allow a model to maintain an uninterrupted linguistic ``memory of thought"—preserving the word-domain KV cache—while the underlying visual context is updated asynchronously. This would move perception closer to an embodied ``flow," where the agent continues to speak or reason without resetting its internal state every time a new visual observation is processed.

\end{adjustwidth}
\end{document}